\documentclass{article}
\usepackage{amsmath} 
\usepackage{graphicx}
\usepackage{booktabs}
\usepackage{multirow}
\usepackage{listings} 
\usepackage{subcaption}
\usepackage{booktabs}
\usepackage{makecell}
\usepackage{caption}
\usepackage{siunitx} 
\usepackage[preprint]{corl_2024} 

\usepackage{listings}

\usepackage{inconsolata}

\lstdefinestyle{txtstyle}{
    basicstyle=\ttfamily\footnotesize\color{black}\linespread{1.1}\selectfont,
    numbers=none,
    numberstyle=\footnotesize,
    stepnumber=1,
    numbersep=8pt,
    showstringspaces=false,
    breaklines=true,
    frame=none,
    backgroundcolor=\color{yellow!6}, 
}
\lstdefinestyle{txtstyle2}{
    basicstyle=\ttfamily\footnotesize\color{black}\linespread{1.1}\selectfont,
    numbers=none,
    numberstyle=\footnotesize,
    stepnumber=1,
    numbersep=8pt,
    showstringspaces=false,
    breaklines=true,
    frame=none,
    backgroundcolor=\color{gray!10}, 
}

\lstdefinestyle{txtstyle3}{
    basicstyle=\ttfamily\footnotesize\color{black}\linespread{1.1}\selectfont,
    numbers=none,
    numberstyle=\footnotesize,
    stepnumber=1,
    numbersep=8pt,
    showstringspaces=false,
    breaklines=true,
    frame=none,
    backgroundcolor=\color{blue!3}, 
}
\usepackage{xcolor}  
\usepackage{fvextra}
\definecolor{verylightgray}{rgb}{0.98, 0.98, 0.98}
\usepackage{listings}

\usepackage{xcolor}  
\usepackage{hyperref} 
\title{AToM-Bot: Embodied Fulfillment of Unspoken Human Needs with Affective Theory of Mind}

\author{
  Wei Ding$^{*1}$, Fanhong Li$^{*1}$, Ziteng Ji$^{2}$, Zhengrong Xue$^{1}$, Jia Liu$^{\dagger1}$ \\
  $^{1}$Tsinghua University \quad $^{2}$University of California Berkeley \\
  \texttt{\{teresading999, lifanhong.ai, zitengjiender0618\}@gmail.com} \\
  \texttt{\{xzr23@mails,liujiaTHU@mail\}.tsinghua.edu.cn}
}
    
\begin{document}
\maketitle
\renewcommand{\thefootnote}{\fnsymbol{footnote}}
\footnotetext[1]{These authors contributed equally to this work.}
\footnotetext[2]{Corresponding author}
\vspace{-0.6cm}
\begin{abstract}
 We propose \textbf{AToM-Bot}, a novel task generation and execution framework for proactive robot-human interaction, which leverages the human mental and physical state inference capabilities of the Vision Language Model (VLM) prompted by the Affective Theory of Mind (AToM). Without requiring explicit commands by humans, AToM-Bot proactively generates and follows feasible tasks to improve general human well-being. When around humans, AToM-Bot first detects current human needs based on inferred human states and observations of the surrounding environment. It then generates tasks to fulfill these needs, taking into account its embodied constraints. We designed 16 daily life scenarios spanning 4 common scenes and tasked the same visual stimulus to 118 human subjects and our robot. We used the similarity between human open-ended answers and robot output, and the human satisfaction scores on a 7-point Likert scale to measure robot performance. AToM-Bot received high human evaluations in need detection ($6.19\pm0.77$), embodied solution ($5.96\pm0.78$), and task execution ($5.98\pm0.75$). We show that AToM-Bot excels in generating and executing feasible plans to fulfill unspoken human needs. Videos and code are available at \href{https://affective-tom-bot.github.io/}{https://affective-tom-bot.github.io/}
\end{abstract}



\begin{figure}[h]
    \centering
    \includegraphics[width=1\textwidth]{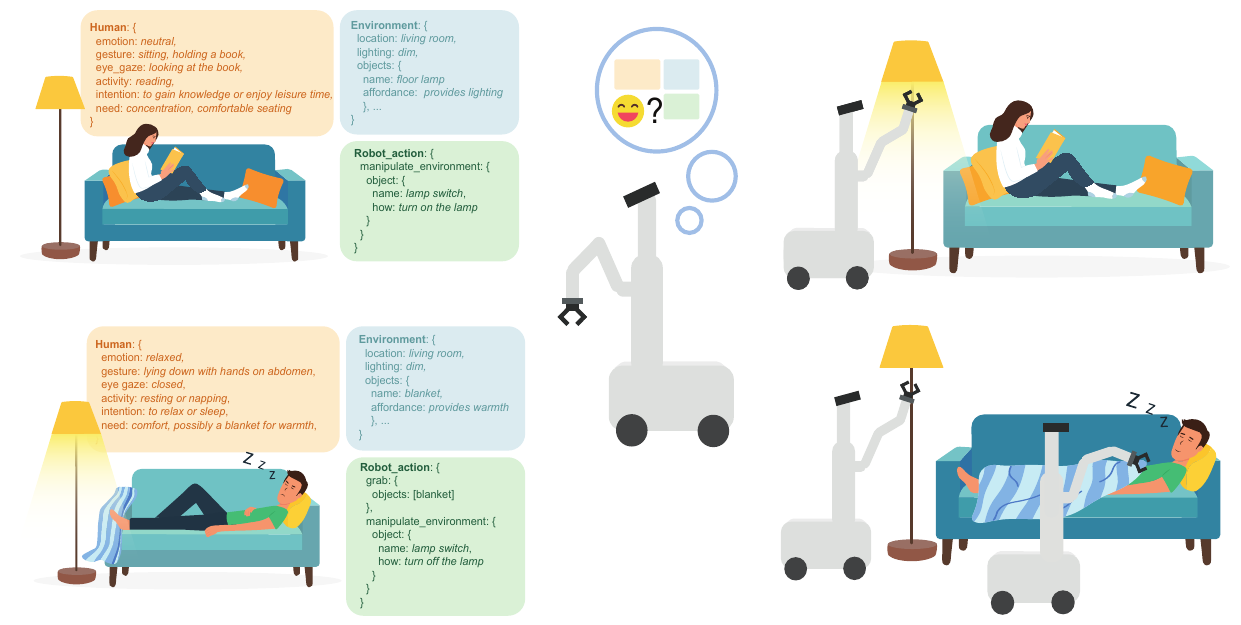}
    \caption{AToM-Bot is a novel task generation and execution framework for proactive robot-human interaction, towards the embodied fulfillment of unspoken human needs.}
    \label{fig:1}
\vspace{-15pt}
\end{figure} 

\section{Introduction}
How can robots fulfill human needs \emph{without being directly told what to do?} Accurately modeling human needs and generating executable plans under physical constraints is a challenge in Human Robot Interaction (HRI). In this work, we studied this challenge by leveraging the vision language model and affective Theory of Mind.
Language is an interactive interface for humans to communicate with robots. There is considerable research on leveraging large language models for robot task planning~\citep{ren_robots_2023, singh_progprompt_2022, huang_language_2022, ren_robots_2023-1, xie_translating_2023, huang_inner_2022} and control grounding~\citep{driess_palm-e_2023,brohan_rt-2_2023,wake_chatgpt_2023,skreta_errors_2023,lin_text2motion_2023,ahn_as_2022,huang_voxposer_2023}. However, most of work begins with a clear verbal command directed at the robot. In real life, many human needs emerge before human formalize them into explicit verbal instructions; for example, when focusing on work, feeling like a cup of tea, we might reach for the tea cup while staring at the computer screen. When we find the tea cup is already empty, we might put it down and continue working rather than pausing work to make a tea. If there's a robot nearby, we may as well not bothering ask it to make tea. But if a robot is proactively observing human, it's not hard for it to infer human current need for ``something to drink'', then the robot can help satisfy the need within its ability. 
There is work focusing on inferring human intention with non-verbal cues with the Theory of Mind(ToM)~\citep{puigWatchAndHelpChallengeSocial2021, cao_smart_2024, zhang_effective_2024, ying_goma_2024}, but not yet grounded on a real-world robot. Such human-agent collaboration tasks require active participation in the same task. Our robot, AToM-Bot, is a robot that understands and observes can use multimodal information to detect human needs and help humans fulfill these non-primary task needs, thus better enabling humans to perform their primary needs. There are three default assumptions of experiment setting. The assumptions in our experiment include:(1) The human in the scenario has at least one need that can be visually identified from a single frame, allowing the robot to take immediate action. (2) The solution to fulfill the need is visible within the scene, eliminating the need for the robot to explore beyond the immediate environment. (3) The needs and solutions are common enough that they do not require additional information specific to the individual. 

Our method proposes \textbf{AToM-Bot} - Affective Theory of Mind robot - a framework for satisfying unspoken human needs utilizing Vision Language Model (VLM)~\citep{openai2023} prompted with Affective Theory of Mind~\citep{shamay2010role, shamay2007dissociable}. We make the following contributions: (1) Given a visual input of a human and their surrounding environment, we use VLM to detect human needs and generate embodied solutions to address these needs. (2) We grounded AToM-Bot with open-vocabulary navigation and imitation learning. (3) We designed 16 common daily scenarios to evaluate robot performance with human satisfaction rating and similarity to human response. AToM-Bot achieved 91.7\%, 87.8\% and 88.1\% satisfaction in need detection, embodied solution and task execution. AToM-Bot represents a new potential for future coexistence and interaction between humans and robots in everyday life.


\section{Related work}
\label{sec:relatedwork}

\subsection{Non-verbal Communication and Robots' Intent Detection}
The integration of large language models (LLMs) into robotics has enhanced robots' ability to interpret human commands through text, speech, and even raised voices~\citep{ren_robots_2023, xie_translating_2023, shi2024yell}. However, a significant portion of human communication is non-verbal, relying on cues such as facial expressions, gaze~\citep{ozdelGazeGuidedGraphNeural2024}, body posture~\citep{luoARBEEAutomatedRecognition2020}, gestures~\citep{yoonRobotsLearnSocial2018}, and physiological signals like heart rate and breathing. These signals often reveal intent before explicit commands are made, enabling robots to anticipate human needs. Recent advancements in multimodal approaches have combined speech, gestures, and gaze~\citep{liu2022intentionbased, manna2022multimodal} to provide a more comprehensive understanding of human intent. However, a key challenge in human-robot interaction (HRI) lies in detecting implicit needs, where intent is weak or unclear. For example, during sleep, subtle cues like changes in posture or breathing may indicate a need for environmental adjustments, such as dimming lights. Most current systems remain reactive, waiting for explicit input. Enhancing robots' ability to detect and respond to these implicit cues is critical for creating more seamless and intuitive human-robot interactions~\citep{puigWatchAndHelpChallengeSocial2021, manna2022multimodal}.

\subsection{Affective Theory of Mind}

Theory of Mind (ToM) refers to the capacity to understand and consider the perceptions, beliefs, and goals of others~\citep{premack1978does, apperly2009humans}.Children as young as three years old demonstrate an initial capability in this cognitive function~\citep{flavell1990developmental, gopnik1992child}. Discussions on ToM for robots began in 2002~\citep{TheoryMindHumanoid}, there have been numerous implementations in simulation and real-world~\citep{wang2024mosaic,caoSmartHelpStrategic2024,patel2022proactive,puig2021watchandhelp}.

In our work, we differentiate between Affective ToM~\citep{li2017comparing} and Cognitive ToM~\citep{shamay2007dissociable}. Cognitive ToM pertains to the ability to comprehend the difference between the speaker’s knowledge and that of the listener, particularly concerning beliefs. Conversely, Affective ToM involves the empathetic recognition of the observed person’s emotional state, focusing on their emotions rather than just their beliefs~\citep{cuccinielloMindPerceptionHRI2023a}. While considerable research has been devoted to robots with Cognitive ToM—spanning applications in plan execution~\citep{devinImplementedTheoryMind2016a}, decision-making~\citep{gorurIntegratingTheoryMind2017}, coordination~\citep{chenVisualBehaviorModelling2021a}, and trust~\citep{patacchiolaDevelopmentalCognitiveArchitecture2022} — Affective ToM has received relatively less attention within robotics. In this context, Affective Theory of Mind enables a robot to empathetically consider the comfort and preferences of a sleeping person, reflecting an understanding of their potential emotional and physical state. The robot effectively places itself in the person’s situation, contemplating what might enhance their comfort. Recent studies have demonstrated that large language models (LLMs) can sometimes perform at or even surpass human levels on specific Theory of Mind tests, such as identifying indirect requests~\citep{strachanTestingTheoryMind2024a}.

\begin{figure}[t]
    \centering
    \includegraphics[width=1\textwidth]{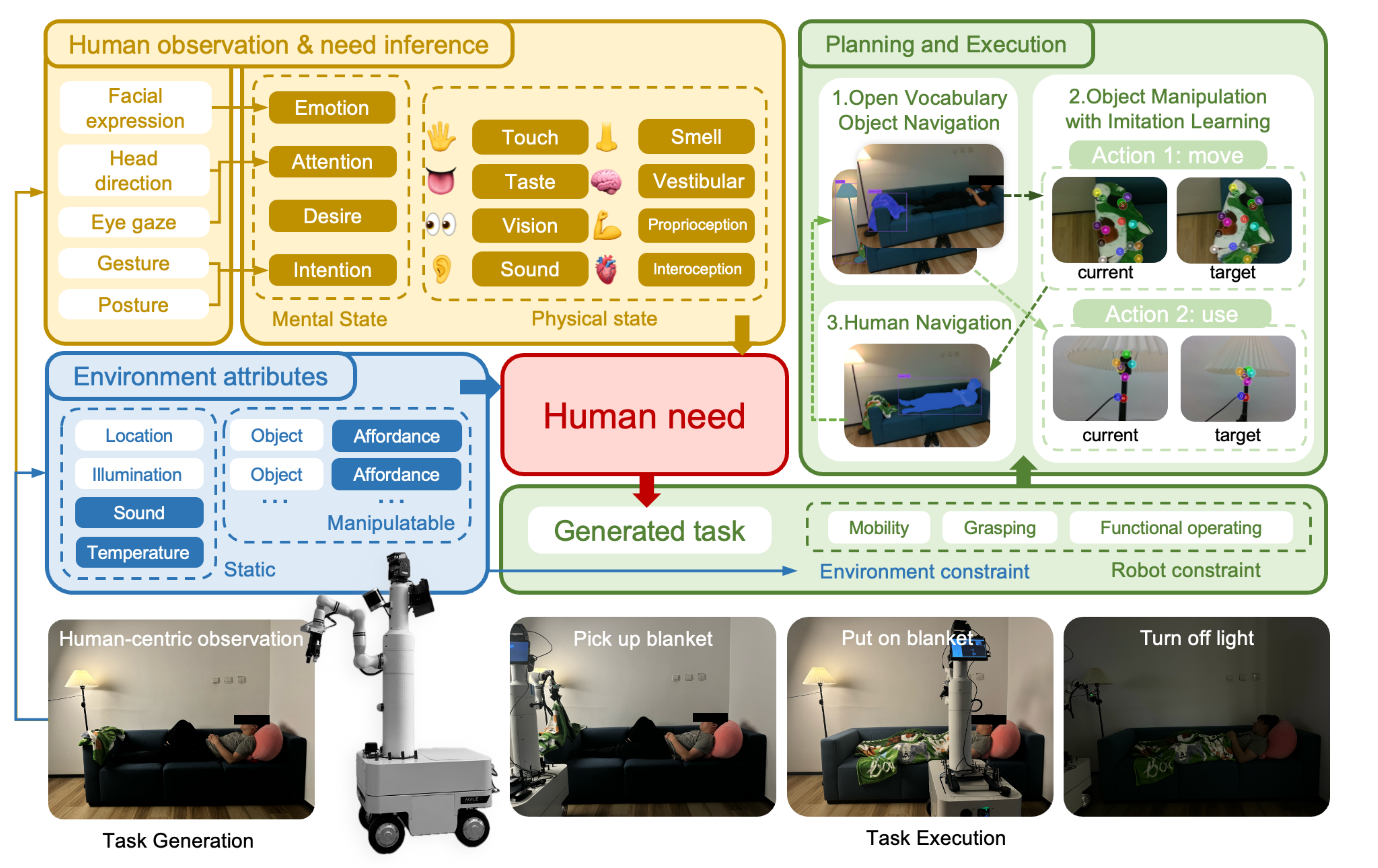}
    \caption{Overview of AToM-Bot, a robotic system for identifying and responding to human needs. It integrates human observations and environmental attributes to infer human needs. It then generates tasks for a robot by navigating to objects, manipulating them, and assisting human in daily setting.}
    \label{fig:fig2}
\end{figure}

\section{Problem statement}

In addressing the challenge of enabling a robot to accurately identify and satisfy human needs, we adopt a comprehensive process that starts with gathering detailed observations and concludes with the executing specified actions. This process includes visual data collection analysis on humans and the environment, inference of human internal states, identification of needs, task formulation, and the final decomposition of tasks into executable actions.

The process begins with the observation $O$ of human behaviors and environmental factors from the input information $I$ (a RGBD image in our case) using a generative pre-trained model $\mathcal{M}_{\text{observe}}$, respectively denoted as: \[
O =\{O_{\text{human}} = \{o_{\text{f}}, o_{\text{e}}, o_{\text{h}}, o_{\text{g}}, o_{\text{p}}\},  O_{\text{environment}} = \{o_{\text{l}}, o_{\text{i}}, \left\{obj_{\text{human}}^{\text{k}}\right\}_{k=1}^{n_{\text{1}}}\}\}=\mathcal{M}_{\text{observe}}(I)
,
\]
where in human observation, \( o_{\text{f}} \), \( o_{\text{e}} \), \( o_{\text{h}} \), \( o_{\text{g}} \), and \( o_{\text{p}} \) denote facial expressions, eye gaze, head direction, gestures, and posture, respectively; in environmental observations, \( o_{\text{l}} \), \( o_{\text{i}} \), \({obj_{\text{human}}^{\text{k}}}\) and \(n_{\text{1}}\) denote location, illumination conditions, interactive object \( k \) for human, the number of interactive object for human respectively.

From these observations, $\mathcal{M}_{\text{state}}$ infers the internal states (\(S\)) of individuals, which includes both mental and physical states such as emotions, intentions, and various sensory inputs:
\[
S = \{S_{\text{physical}} = \{s_{\text{t}}, s_{\text{ta}}, s_{\text{v}}, s_{\text{s}}, s_{\text{sm}},s_{\text{ve}},  s_{\text{p}}, s_{\text{i}}\}, S_{\text{mental}}=\{s_{\text{em}}, s_{\text{a}}, s_{\text{d}}, s_{\text{in}}\}\}= \mathcal{M}_{\text{state}}(O),
\]
where in physical states, \( s_{\text{t}} \), \( s_{\text{ta}} \), \( s_{\text{v}} \), \( s_{\text{s}} \), \( s_{\text{sm}} \), \( s_{\text{ve}} \), \( s_{\text{p}} \), and \( s_{\text{i}} \) denote touch, taste, vision, sound, smell, vestibular, proprioception, and interoception respectively; in the mental states, \( s_{\text{em}} \), \( s_{\text{a}} \), \( s_{\text{d}} \), and \( s_{\text{in}} \) denote emotion, attention, desire, and intention respectively.

These states allow us to identify any unmet needs (\(N\)): \( N = \mathcal{M}_{\text{needs}}(S) \). Upon identifying these needs, we analyze the need and constraint  (\(C\)) of the robot body and the environment to generate feasible tasks (\(T\)) by$ \mathcal{M}_{\text{task}}$, denoted as: 
\[
T = \mathcal{M}_{\text{task}}\{N, C=\{{c_{\text{m}}, c_{\text{g}}, c_{\text{o}}}, \left\{obj_{\text{robot}}^{\text{k}}\right\}_{k=1}^{n_{\text{2}}}, \left\{obj_{\text{robot}}^{\text{potential, k}}\right\}_{k=1}^{n_{\text{3}}}\}\},
\]
where in the robot's body constraints, \(c_{\text{m}} \), \(c_{\text{g}} \) and \(c_{\text{o}} \) represent mobility, grasp ability and functional operating ability, respectively; and in the the environmental context \({obj_{\text{robot}}^{\text{k}}}\) represents the visible and manipulable objects for robots, \({obj_{\text{robot}}^{\text{potential, k}}}\) represents the potential existing objects for robots in this specific environment.

Finally, the formulated tasks are decomposed into a sequence of $t$ steps executable robotic actions (\(A\)) by $ \mathcal{M}_{\text{decompose}}$:
\[
A = \mathcal{M}_{\text{decompose}}(T)=\{a_1, a_2, \ldots, a_t\}, a_t \subseteq \{a_m, a_g, a_o\},
\]where  $a_m$, $a_g$, $a_o$ denotes moving to a certain target(object or human), grasping a certain object and operating a certain object.

\section{Method}
We present a robot-human interaction framework designed to proactively detect and address human needs empowered by open-vocabulary perception and deduction capabilities of VLM and visual foundation model (See Figure. ~\ref{fig:fig3}). 
To be more specific, we employ GPT-4V~\citep{gpt4v} to (i) extract the observation of human behaviors and environment factors, (ii) infer internal human states from the observations, and (iii) finally generate executable tasks catering to the unexpressed human needs. As for the physical execution of the generated task plan, we rely on Grounding SAM~\citep{kirillov2023segany,liu2023grounding,ren2024grounded} to pinpoint the locations of the concerning objects and human users, and DINO-ViT features ~\citep{amir2021deep,dipalo2024dinobot}to retrieve the knowledge of low-level manipulation skills from pre-recorded human demonstrations.


\subsection{Proactive Human Need Detection}
Our robot employs a head-mounted camera to observe humans. Relying solely on raw observations for input into VLM often leads to misinterpretations of human states. \emph{How do human infer another person's internal states based solely on image?}
In fact, when we engage in affective social reasoning, it is easy to re-experience and bring past experiences into the process~{\citep{perner2007episodic,gallese2007before}}. Recalling these experiences can also activate related sensory pathways, causing similar responses. This capacity for empathetic thinking is precisely what robots lack. Here, we employ reasoning based on common sense, using typical human mental states \citep{beaudoin2020systematic,anderson2004integrated} and physical states \citep{goldstein2002sensation,ritchie2015bodily} as cues to prompt robots to deduce current human feelings through visual clues. 

From a first-person viewpoint, the camera captures various non-verbal human expressions such as facial expressions, eye gaze, head direction, gestures, and posture, along with environmental attributes like location, lighting, and nearby objects. These observations enable us to deduce mental states (e.g., emotions, attention, intentions) and physical conditions (e.g., senses of touch, taste, vision, hearing, smell, as well as vestibular, proprioceptive, and interoceptive senses), and also to recognize environmental factors beyond visual cues, such as sound and temperature.

Integrating all these elements, we propose a structured prompt for the VLM that systematically incorporates these diverse data streams to infer human needs, as depicted in Figure 2. Through a detailed description of relative factors, we bring the subconscious AToM in humans during social inference into our robot, which greatly enhances its ability to understand human internal states.

\subsection{Embodied Task Generation for Human Need Solution}
Upon identifying a specific human need (or potentially multiple needs), the robot should devise a practical and immediate solution. Equipped with mobility and the ability to manipulate objects with a single hand, the robot's actions must be feasible and directly executable \citep{ren_robots_2023,ahn_as_2022}. Instead of complex tasks like completing someone's work or moving heavy appliances like a fridge, the robot should focus on simpler, achievable tasks within its capabilities and current context.

The robot can interact with objects visible in the scene or infer the presence of everyday items based on the scene, utilizing common sense reasoning. For example, it might predict a yoga block in a gym but not in an office. The robot's manipulative abilities are categorized into two primary actions: moving and using objects. This distinction is crucial due to its single gripper and one-arm configuration.

The VLM evaluates whether fulfilling a need involves merely moving or actively using an object. For instance, transferring a blanket to a person might satisfy a need, whereas merely moving a lamp to the person would be inappropriate and ineffective. If a need involves using an object, the robot must determine whether it can perform the task given its configuration. For example, it can turn on a light but cannot operate a massage gun for a human. This approach ensures that the robot's responses are both relevant and within its operational scope.

\subsection{Open-Vocabulary Mobile Manipulation for Task Execution}

\textbf{Task-Conditioned Navigation} Once the robot generates a task to perform, it employs its pan-tilt unit (PTU) to scan the environment dynamically. The scan starts from a central, neutral position and includes systematic horizontal and vertical sweeps, ensuring comprehensive visual coverage. The system uses the Grounding SAM~\citep{ren2024grounded} algorithm to segment and identify relevant objects from the visual data. Upon detecting an object that matches the specified criteria, the robot utilizes depth information from RGB-D images to pinpoint the object's location.

Once the position of the object is calculated, the robot plans its navigation, balancing the necessity of proximity for precise manipulation with the importance of maintaining a safe distance. This balance is achieved by analyzing potential manipulation trajectories, whether horizontal or vertical. The robot then positions itself into a pre-manipulation pose tailored for the end effector. The navigation goal is set at a point along the extended line of the end effector’s z-axis, optimizing the approach towards the object. This methodical planning ensures the robot can efficiently and safely interact with its environment to accomplish the task.

Additionally, the robot must locate the person, if the identified task involves moving an object to a human. For this, we utilize the Grounding SAM~\citep{ren2024grounded} algorithm to detect humans within the environment. The system then computes the similarity between the currently detected masked image of the human and the one initially captured to ensure that the robot approaches the correct individual.

\textbf{Manipulation via Trajectory Alignment} For object manipulation, we follow the DINOBot~\citep{dipalo2024dinobot} framework, which utilizes DINO-ViT features to perform the spatial alignment between the current observation and the pre-recorded demonstrations for manipulation skills.
When the robot encounters an object, it uses image-level semantic features to identify the most visually similar object from a database of historical human demonstrations. Once a visually similar object is identified, the robot utilizes pixel-level geometric features to precisely align the relative spatial position and orientation between the camera frame and the target object. This alignment, crucial for effective interaction, involves adjusting the robot’s position and orientation to align with the pre-recorded trajectory. Once there is an error between the current relative pose and the pre-recorded relative pose, the robot arm executes a delta end-effector path to complete the manipulation task.

\subsection{Analysis of Human and Robot Responses}

\noindent To evaluate the alignment of robot-generated responses with human descriptions, we employed a clustering-based approach followed by cosine similarity and proportional analysis. The method is outlined as follows:

\paragraph{Data Preprocessing}
The responses from the open vocabulary survey were provided in Chinese. To facilitate further analysis, the following preprocessing steps were applied: Each response was segmented into units based on punctuation since participants were asked to include several needs or solutions in response to one image. The segmented text units were translated from Chinese to English using the Helsinki-NLP's MarianMT model \texttt{opus-mt-zh-en}.

\paragraph{Clustering of Human responses} 
For open-vocabulary human-generated needs and solutions, text embeddings were then generated for these descriptions, We utilized the Sentence-BERT model \texttt{paraphrase-} \texttt{multilingual-} \texttt{MiniLM-L12-v2} to encode both human and robot responses into dense embeddings for semantic analysis. The responses were grouped into clusters $C = \{C_1, C_2, \ldots, C_k\}$, where each cluster $C_i$ contains descriptions with similar content. The optimal number of clusters, $k$, was initially determined using the elbow method and further adjusted based on the clustering results.

\paragraph{Cosine Similarity}
To quantify how closely a robot-generated response $r_j$ aligns with human responses, we compute the cosine similarity between $r_j$ and each human response $h_i$ within the same cluster $C_{i^*}$ to which $r_j$ has been assigned. The average cosine similarity for the robot response $r_j$ is given by:

\begin{equation}
\text{AvgSim}(r_j) = \frac{1}{|C_{i^*}|} \sum_{h_i \in C_{i^*}} \text{cos\_sim}(r_j, h_i)
\end{equation}

where $\text{cos\_sim}(r_j, h_i)$ represents the cosine similarity between the robot response $r_j$ and a human response $h_i$ within cluster $C_{i^*}$, and $|C_{i^*}|$ is the number of human responses in that cluster. This metric provides a measure of the semantic similarity between the robot and human responses.

\paragraph{Proportion Rate:}
The proportion rate $P_{i^*}$ represents the fraction of total human responses that belong to the same cluster $C_{i^*}$ as the robot response $r_j$. It is calculated as follows:

\begin{equation}
P_{i^*} = \frac{|C_{i^*}|}{n}
\end{equation}

where $|C_{i^*}|$ is the number of human responses in cluster $C_{i^*}$, and $n$ is the total number of human responses across all clusters. The proportion rate indicates how representative the cluster containing the robot response is of the entire dataset, with higher values suggesting that the robot's response is aligned with a common human perspective.

\begin{figure}[h]
    \centering
    \includegraphics[width=1\textwidth]{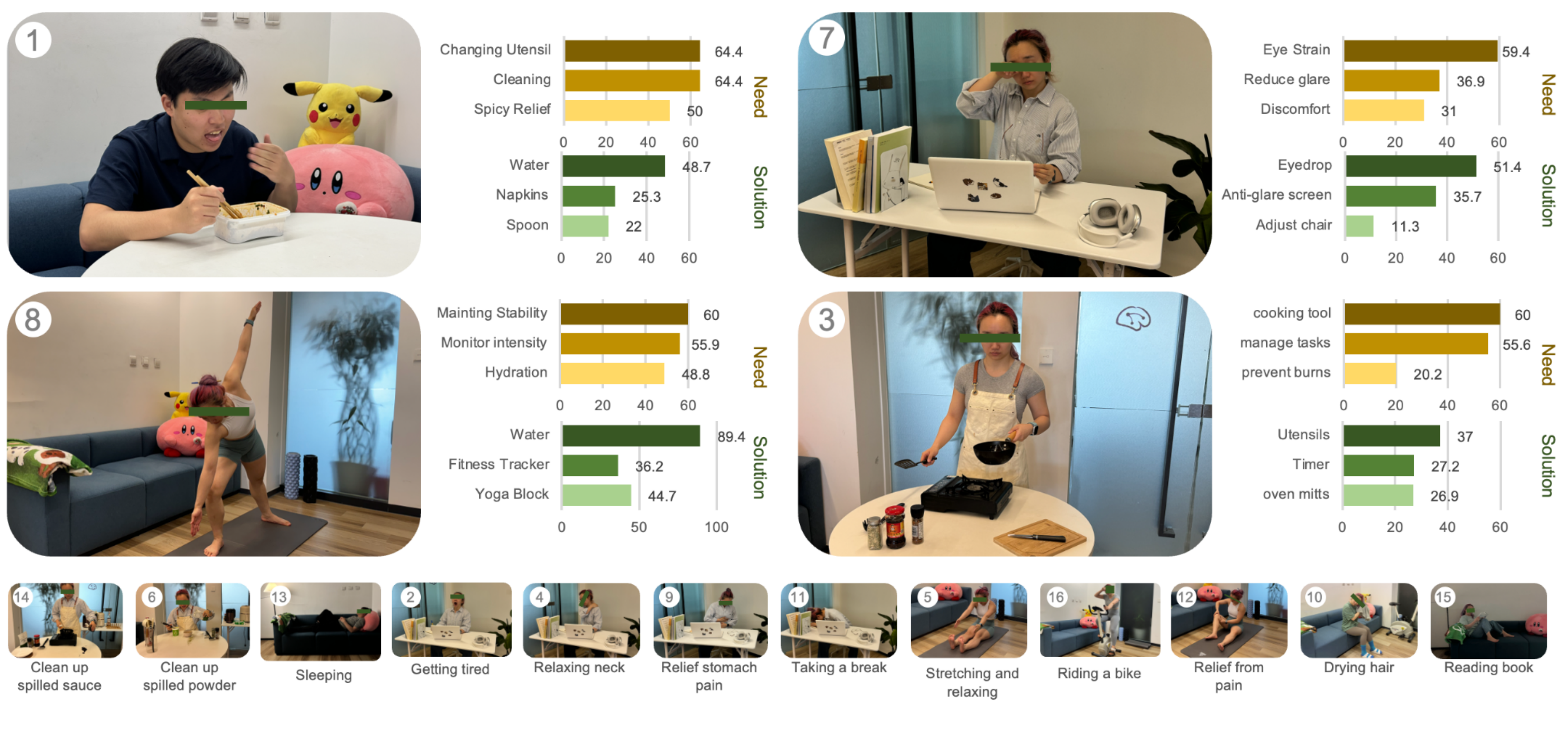}
    \caption{Examples of needs and solutions generated by AToM-Bot for selected task scenarios. The bar graphs display the proportion of responses indicating perceived needs (brown) and potential solutions (green) for specific daily activities, including eating spicy food (1), feeling tired while working (7), doing balance yoga exercises (8), and cooking (3). Each task scenario shows the responses from 118 participants, where each participant could provide one or more responses. The displayed proportions are calculated as the percentage of each response type relative to the total number of participants (118). The displayed proportions are calculated as the percentage of each response type relative to the total number of participants (118). Additional 12 tasks are shown at the bottom.}
    \label{fig:fig3}
\end{figure}

\section{Experiments}

\subsection{Experimental setup}
\textbf{Scenarios: }Our experiment setup involved 16 scenarios (Figure. \ref{fig:fig3}) conducted within four commonly encountered everyday environments: office spaces, home gyms, living rooms, and kitchens. In each setting, humans typically encounter various physical or mental states of distress. We displayed 4 typical instances from each environment on the upper half: eating spicy food in the kitchen, feeling tired in the office, sweating while cycling in a home gym, and falling asleep on a sofa. In order to provide more possibilities for the robot to operate and ensure the ecological validity of the test, we also prepared objects matching the scene in the same space outside the picture, ranging from 10 to 18 kinds for each scene. This includes the solution for the robot we tested in advance in Section 5.3.

Our experiments and robot demonstrations are designed with simplified task settings to validate our concept. The key assumptions include:

\begin{enumerate}
    \item The human in the scenario has at least one visually identifiable need from a single frame, enabling the robot to take immediate action.
    \item The solution to fulfill the need is visible within the scene, so the robot does not need to explore beyond the immediate environment.
    \item The needs and solutions are common enough that they do not require individual-specific information.
\end{enumerate}

\textbf{Human-centered Metrics: }We deployed our robot to react to these scenarios and recruited 118 participants to conduct a series of evaluations on satisfaction with various stages of our system and its similarity to the human mind. When combining both groups, the overall mean age was 23.54 years (SD = 2.90), with ages spanning from 18 to 33 years. The total sample consisted of 78 females (66.1\% of the total sample) and 40 males (33.9\%).

\textbf{Robot Setup: }The robot is a mobile manipulator with a 7 degree-of-freedom arm from Realman, a parallel gripper from DH-ROBOTICS, and an omnidirectional mobile base from Agilex. It uses two Realsense 435 cameras, with the top camera running at 640x480, and the wrist camera running at 320x240.

\subsection{Comparison and Evaluation}
\begin{table}[t]
\centering
\scriptsize
\setlength{\tabcolsep}{3pt}
\begin{tabular}{@{}c ccc ccc cc@{}}
\toprule
\multirow{2}{*}{\makecell{Task \\ ID}} & \multicolumn{3}{c}{Need Detection}      & \multicolumn{3}{c}{Embodied Solution}     & \multicolumn{2}{c}{Task Execution} \\
\cmidrule(lr){2-4}                      \cmidrule(lr){5-7}                        \cmidrule(l){8-9}
                                       & Similarity & Proportion Rate & Satisfaction & Similarity & Proportion Rate & Satisfaction & Success Rate & Satisfaction \\ 
\midrule
1  &  $39.4\%$  & $64.4\%$  & $6.59\pm0.62$    & $48.7\%$   & $73.7\%$  & $6.23\pm0.98$    & $8/10$ & $6.46\pm0.72$ \\
2  &  $49.1\%$  & $45.7\%$  & $6.27\pm0.86$    & \pmb{$66.3\%$}   & $49.2\%$  & $5.42\pm1.64$    & $9/10$ & $5.83\pm1.33$ \\
3  &  $60.0\%$  & $37.2\%$  & $5.83\pm1.15$    & $37.0\%$   & $68.6\%$  & $5.83\pm1.42$    & $5/10$ & $6.25\pm0.93$ \\
4  &  $64.8\%$  & $48.3\%$  & $6.58\pm0.62$    &  $55.8\%$  & $21.7\%$  & $6.63\pm0.51$    & $8/10$ & $6.24\pm1.07$ \\
5  &  $75.2\%$  & $39.0\%$  & $6.55\pm0.69$    &  $40.8\%$  & $50.8\%$  & $6.52\pm0.71$    & $4/10$ & $6.52\pm0.83$ \\
6  &  $34.1\%$  & $57.6\%$  & $6.63\pm0.68$    &  $58.4\%$  & $50.0\%$  & $5.39\pm1.97$    & $10/10$ & $5.52\pm1.66$ \\
7  &  $59.4\%$  & $72.8\%$  & $6.61\pm0.57$    & $51.4\%$  & $65.3\%$  & $6.59\pm0.69$    & $4/10$ & $6.42\pm0.91$ \\
8  &  $59.5\%$  & $35.6\%$  & $5.79\pm1.47$    & $44.7\%$  & $60.1\%$  & $5.54\pm1.56$    & $10/10$ & $6.31\pm1.14$ \\
9  &  $62.7\%$  & $61.8\%$  & $6.45\pm0.77$    & $36.4\%$  & $49.2\%$  & $6.52\pm0.74$    & $5/10$ & $6.11\pm1.28$ \\
10 &  $56.3\%$  & \pmb{$100\%$}  & $6.68\pm0.55$    & $49.5\%$  & $75.4\%$  & \pmb{$6.76\pm0.49$}    & $10/10$ & $6.79\pm0.48$ \\
11 &  $65.7\%$  & $38.9\%$  & $6.55\pm0.77$    & $38.1\%$   & $66.1\%$  & $5.37\pm1.91$    & $5/10$ & $4.96\pm1.82$ \\
12 &  $69.6\%$  & $83.1\%$  & $6.62\pm0.59$    & $29.0\%$  & $53.4\%$  & $6.65\pm0.70$    & $9/10$ & $6.42\pm0.94$ \\
13 &  $73.2\%$  & $55.1\%$  & $6.48\pm0.86$    & $54.8\%$  & $80.5\%$  & $6.37\pm1.12$    & $4/10$ & \pmb{$6.82\pm0.52$} \\
14 &  $38.6\%$  & $52.7\%$  & $6.66\pm0.58$    & $60.2\%$  & $48.3\%$  & $6.55\pm0.83$    & $7/10$ & $6.34\pm0.92$ \\
15 &  $61.4\%$  & $44.9\%$  & \pmb{$6.70\pm0.61$}    & $55.8\%$  & \pmb{$91.5\%$}  & $6.69\pm0.68$    & $4/10$ & $6.55\pm0.75$ \\
16 &  \pmb{$78.8\%$}   & $73.7\%$  & $5.82\pm1.40$    & $57.1\%$  & $44.9\%$  & $5.78\pm1.68$    & $9/10$ & $5.91\pm1.55$ \\
\midrule
\textbf{Average} &  $59.2\%$  & $56.9\%$ & $6.42\,/\,7$ &  $49.0\%$ & $59.3\%$ & $6.15\,/\,7$ & $69.4\%$ & $6.17\,/\,7$ \\
\bottomrule
\end{tabular}
\vspace{5pt}
\caption{Combined Human Comparison and Satisfaction Evaluation of Need Detection, Embodied Solution, and Task Execution of AToM-Bot.}
\vspace{-15pt}
\end{table}

\textbf{Human Need Detection} In Experiment 1, participants were presented with images depicting various scenarios and asked to provide open-ended responses identifying the inferred needs. Subsequently, they evaluated the robot’s responses (output of using the prompt described in Section 4.1) on a 7-point Likert scale, where 1 indicates extreme dissatisfaction and 7 indicates extreme satisfaction. We employed text embedding and unsupervised clustering algorithm \citep{guo2003knn} to categorize the responses into 3-5 distinct needs and corresponding solutions. The need most frequently identified by humans served as the standard for comparing the robot's generated needs, resulting in a Similarity score. The average similarity score was 72.8\%. The overall satisfaction rating was $6.42 \pm 0.32$ out of 7, suggesting that, despite variability in human output, participants were generally delighted with our outputs, and the robot's output is very similar to the average human. The human need detection phase demonstrated a high degree of alignment between the robot’s inference of needs and human perceptions.

\textbf{Embodied Task Generation as Human Need Solution} In Experiment 2, participants were asked to generate solutions corresponding to the needs in Experiment 1 and to rate satisfaction with the solutions generated by the robot. The analysis methods were the same as those used in Experiment 1. The average similarity score was 69.6\%. The overall satisfaction rating was $6.15 \pm 0.55$. It is noteworthy that in Tasks 5 and 8, the similarity between the robot's and the human's solutions was relatively low (16.9\% and  13.1\%), yet the satisfaction ratings remained high( $6.52 \pm 0.71$ and  $5.54 \pm 1.56$ ). This is because most humans chose to directly use their hands to help stabilize or push the participant for better stretching, an approach unsuitable for our robot. Instead, our robot opted to find assisting equipment within its capability range, such as yoga bricks or resistance bands. This is a prime example of the effectiveness of the embodied constraints in our prompt.

\textbf{Real World Execution} For real-world implementation, we recorded videos of the robot successfully executing solutions from Experiment 2 in a third-person perspective. Participants scored the keyframes for satisfaction, resulting in an average satisfaction rating of $6.17 \pm 0.54$. We also evaluated the robot's task success rate. Since the tasks were long-horizon, involving various factors such as object detection accuracy, spatial alignment precision with records, and navigation execution accuracy, there were considerable fluctuations in success rates, especially in tasks requiring precise operations like turning on lights or handling soft, whole-body movements like covering with a blanket. Nevertheless, the overall performance was $69.4 \% $. See details in \href{https://affective-tom-bot.github.io/}{AToM-Bot}.



\subsection{Ablations}
We performed ablation studies to assess the impact of key components: AToM for need detection and the embodied constraints for grounding task generation in our prompt. These components were evaluated for their contribution to need similarity (Table~\ref{table-need-similarity}) and solution similarity (Table~\ref{table-solution-similarity}) , detailed in Appendix. \textbf{AToM:} Removing AToM significantly lowers need and solution similarity by 26.4\% and 31.2\%, respectively. \textbf{Embodied Constraints:} Operating without constraints results in much lower scores in both need and solution similarity by 4.1\% and 38.6\%. Employing both AToM and embodied constraints as AToM-bot achieves the highest improvements in need and solution similarity, reaching 72.8\% and 69.6\%. These results confirm that both AToM and embodied constraints are essential for aligning with human reaction. The ablation studies further solidified the importance of both AToM for accurate need detection and the role of embodied constraints in ensuring the practicality of generated tasks.

\begin{table}[h]
\centering
\small 
\begin{minipage}{0.45\linewidth}
    \centering
    \begin{tabular}{ccc}
        \toprule
        Prompt     & w/o AToM     & w/ AToM \\
        \midrule
        w/o Constraints &$33.2\% $ & $68.7\% $ \\
        w/ Constraints     & $46.4\% $&  \pmb{$72.8\% $}     \\
        \bottomrule
    \end{tabular}
    \vspace{5pt}
    \caption{Need similarity.}
    \label{table-need-similarity}
    \vspace{-0.3cm}
\end{minipage}
\hspace{0.05\linewidth}
\begin{minipage}{0.45\linewidth}
    \centering
    \begin{tabular}{ccc}
       \toprule
        Prompt     & w/o AToM     & w/ AToM \\
        \midrule
        w/o Constraints & $4.9\% $&  $31.0\% $   \\
        w/ Constraints     & $38.4\% $ &  \pmb{$69.6\% $}    \\
        \bottomrule
    \end{tabular}
    \vspace{5pt}
    \caption{Solution similarity.}
    \label{table-solution-similarity}
    \vspace{-0.3cm}
\end{minipage}
\end{table}

These results highlight AToM-Bot’s capability to not only understand and predict human needs effectively but also to generate and execute tasks that address these needs in a real-world setting.


\section{Limitation and Future Work}
Despite compelling results, AToM-Bot has several limitations. First, the robot is equipped with only two cameras, limiting its sensory input to visual information. In future enhancements, the robot could be equipped with multi-modal sensors like audio sensors and thermometers, enabling it to receive more information and analyze human needs with greater breadth.

Perceptual errors stem from limited information. Humans infer mental states from observed actions, even with minimal cues. Theory of Mind (ToM) involves inferring unseen cognitive states from limited data. As more information becomes available, perceptual errors decrease, and enriched task settings will provide additional context for the robot to better infer human needs. 
incorporate with sensors for human biological signals \citep{yang_artificial_2022} will assist in the early diagnosis of medical conditions. 
Second, the current experiment did not employ sequential detection; long-term dynamic observations rather a snapshot will further increase the detection precision. Third, all the target objects are within the robot's visibility by turning the PTU. If equipped with zero-shot object navigation modules \citep{cow,vlmap} the robot will have the potential to navigate outside the room for more complex task and more possible solutions. Fourth, we observed cases where human participants' solutions differed from those of the robot in solutions to eating spicy food and stomach discomfort that predominantly resulted from cultural backgrounds. Future involvement of participants from diverse cultural backgrounds will provide greater diversity. Lastly, although our framework exhibits a certain degree of generalizability, where similar visual features of objects allow for the transfer of recorded trajectories—such as among different cups or between handheld vacuum and hair dryers—these capabilities are still constrained by our reliance on pre-recorded movements. 

In the future, we could enhance the robot's flexibility by exploring more trained end-to-end models that extend from visual features to low-level actions, thereby developing more adaptable and autonomous movement strategies. In future work, we plan to enhance the complexity of each assumption to better reflect real-world scenarios: Scenarios where the human has no discernible needs or where no immediate actions can be taken by the robot. Here, the robot would dynamically infer the need from video or multiple perceptions and determine optimal action timing. Situations where the robot needs to explore outside the initial environment to find a solution or where the planned solution is unavailable, requiring creative problem-solving. Personalization of needs through long-term interaction with humans. These steps will enrich the AToM bot task set. Although our current task setting is simplified, it effectively demonstrates that a robot can infer human needs and provide assistance without explicit commands. The primary goal of the AToM-Bot is to design and validate a framework capable of inferring human mental and physical states, and generating appropriate tasks based on these inferences. As a result, the initial tasks designed for the AToM-Bot are relatively basic, covering a range of foundational scenarios. Future work includes handling more complex tasks, as well as provide personalized, adaptive solutions tailored to the unique needs and preferences of individual users. This progression will move the AToM-Bot closer to practical real-world applications, with greater utility and customization for diverse user requirements.


\section{Conclusion}

In this work, we introduce \textbf{AToM-Bot}, a comprehensive framework for inferring human mental and physical needs, utilizing Vision Language Models enhanced with Affective Theory of Mind. AToM-Bot proactively generates and executes tasks based on these inferences, tailored to real-world environments and everyday scenarios. This approach provides substantial benefits in adapting to a wide range of unspoken instructions and dynamic human contexts, enabling more intuitive and effective robot-human interactions.

By successfully deciphering unspoken human needs and generating actionable solutions within the constraints of the real world, AToM-Bot not only demonstrates its capabilities but also its significant potential for revolutionizing human-robot interaction. These findings not only open the door for future enhancements but also herald a new era of broader applications for robotic systems in everyday human environments, inspiring optimism for the future of this field.


\clearpage


\bibliography{example}  

\newpage
\appendix
\section{Appendix}


\subsection{Prompts}

\subsubsection{AToM-Bot Prompt}
\begin{lstlisting}[style=txtstyle]
You are a robot with one robot arm(not enough to reach the ground, so that you can not pick up anything that's on the ground) and a mobile base, and you have no hearing and verbal modules; you have high emotional intelligence, designed to analyze human needs and give viable solutions for better human experiences. You will provide detailed JSON outputs that include all identified attributes and suggested interventions to help ensure effective and timely support. You identify emotions, gestures, and actions and recommend specific supportive actions based on these observations. This includes suggestions for how robots or automated systems can assist individuals in the image. Please carefully examine the difficulty that the person in the image faces, which means their facial expression, posture, eye gaze, head direction, and gestures are significant when it comes to determining the problem that the person is facing. As a result, please pay extra attention and carefully examine them before outputting solutions. It is ideal to find a solution that only interacts with objects rather than humans. Your solution should be feasible and possibly use objects visible in the picture, and there has to be an action in the solution. Since you have a moving base, you can look around and navigate in the environment that you are currently in for objects that might help the person. Also note that verbal suggestions and tips are not accepted, and you cannot interact with computers, phones, and laptops. In addition, output some items that are not in the current environment but might help the person in the current activity; it is ok to provide medicine when the person in the image is experiencing pain. Please carefully examine these items before outputting the solution, and these items should play an essential role in your decision-making (or possibly be part of your solution). I need 6 needs in pairs with 6 solutions in total, and 3 pairs of them should use the possible items. JSON format example\: \{Environment\: \{location\: , lighting\: ,sound\: ,temperature\: , objects\: \{name\: affordance\}, possible item\: \{items:\}, human\: \{gesture\: , eye gaze\: , activity\: , emotion\: , intention\: , touch\: , taste\: , vision\: , smell \: , sound\: , vestibular\: , proprioception\: , interoception\: ,needs\: \{need1\: \{description\: , suggested robot solution\: \}\}\}\}
\end{lstlisting}

\subsubsection{Action Generation Prompt}
\begin{lstlisting}[style=txtstyle3]
You are a robot that translates sentences of solutions to robot action. Your constraint is that you only have one arm and a moving base. I will give you multiple solution requirements, you will output the robot action in JSON format (select if you want to grab the object or manipulate the object). JSON format example: {solution requirement 1: {navigation(object_name): , select between move(object_name,target_location(e.g. person)) or use(object_name):}}. You should notice that 'move' means take the object to a specific location, e.g. fetch a chair, while 'use' means you use it according to its functional purpose, e.g. sit on a chair.
\end{lstlisting}

\newpage
\subsection{Need Detection, Embodied Solution and Action List Output}
\subsubsection{Task 1}
\textbf{Need Detection and Embodied Solution}
\begin{lstlisting}[style=txtstyle]
{
    "Environment": {
        "location": "Indoor dining area",
        "lighting": "Bright artificial light",
        "sound": "Quiet",
        "temperature": "Comfortable",
        "objects": {
            "table": "eating surface",
            "food container": "holds food",
            "chopsticks": "eating utensil",
            "plush toys": "decorative"
        },
        "possible item": {
            "items": [
                "water bottle",
                "napkins"
            ]
        },
        "human": {
            "gesture": "hand near mouth, holding chopsticks",
            "eye gaze": "looking at food",
            "activity": "eating",
            "emotion": "discomfort",
            "intention": "to eat",
            "touch": "active",
            "taste": "active",
            "vision": "active",
            "smell": "active",
            "sound": "inactive",
            "vestibular": "inactive",
            "proprioception": "active",
            "interoception": "active",
            "needs": {
                "need1": {
                    "description": "Person appears to be experiencing discomfort possibly due to spicy or hot food",
                    "suggested robot solution": "Retrieve a water bottle from nearby and place it within easy reach of the person to help alleviate discomfort."
                },
                "need2": {
                    "description": "Potential mess from eating, as indicated by the person's focused handling of food",
                    "suggested robot solution": "Provide napkins by picking them up from a nearby dispenser and placing them on the table."
                },
                "need3": {
                    "description": "Person might need a change in eating utensil if discomfort continues",
                    "suggested robot solution": "Locate a spoon nearby and replace the chopsticks with it to facilitate easier eating."
                }
            }
        }
    }
}
\end{lstlisting}

\textbf{Action List}
\begin{lstlisting}[style=txtstyle3]
{
    "Retrieve a water bottle from nearby and place it within easy reach of the person to help alleviate discomfort": {
        "navigation": "water bottle",
        "move": "water bottle, person"
    },
    "Provide napkins by picking them up from a nearby dispenser and placing them on the table": {
        "navigation": "napkin dispenser",
        "move": "napkins, table"
    },
    "Locate a spoon nearby and replace the chopsticks with it to facilitate easier eating": {
        "navigation": "spoon",
        "move": "spoon, chopsticks"
    }
}
\end{lstlisting}

\subsubsection{Task 2}
\textbf{Need Detection and Embodied Solution}
\begin{lstlisting}[style=txtstyle]
{
    "Environment": {
        "location": "office",
        "lighting": "bright",
        "sound": "quiet",
        "temperature": "comfortable",
        "objects": {
            "laptop": "work tool",
            "headphones": "audio device",
            "books": "information source",
            "pen": "writing tool",
            "notebook": "note-taking"
        },
        "possible_items": {
            "items": [
                "coffee maker",
                "water dispenser",
                "ergonomic chair"
            ]
        }
    },
    "human": {
        "gesture": "yawning",
        "eye_gaze": "forward",
        "activity": "working on laptop",
        "emotion": "tired",
        "intention": "to continue working",
        "touch": "contact with laptop",
        "taste": "none",
        "vision": "focused on screen",
        "smell": "none",
        "sound": "none",
        "vestibular": "sitting",
        "proprioception": "aware of body posture",
        "interoception": "feeling of fatigue",
        "needs": {
            "need1": {
                "description": "Reduce fatigue and increase alertness",
                "suggested_robot_action": "navigate to a nearby coffee maker, retrieve a beverage, and deliver it to the person"
            },
            "need2": {
                "description": "Improve comfort to enhance focus and reduce strain",
                "suggested_robot_action": "Retrieve an ergonomic chair from the office storage and replace the current chair to enhance comfort and posture."
            },
            "need3": {
                "description": "Maintain engagement and productivity",
                "suggested_robot_action": "Position a desk lamp near the workspace to provide better lighting conditions, especially beneficial during long working hours."
            }
        }
    }
}
\end{lstlisting}

\textbf{Action List}
\begin{lstlisting}[style=txtstyle3]
{
    "navigate to a nearby coffee maker, retrieve a coffee, and deliver it to the person": {
        "navigation": "coffee maker",
        "move": "coffee, person"
    },
    "Retrieve an ergonomic chair from the office storage and replace the current chair to enhance comfort and posture.": {
        "navigation": "office storage",
        "move": "ergonomic chair,desk"
    },
    "Position a desk lamp near the workspace to provide better lighting conditions, especially beneficial during long working hours.": {
        "navigation": "desk lamp",
        "move": "desk lamp, workspace"
    }
}
\end{lstlisting}

\subsubsection{Task 3}
\textbf{Need Detection and Embodied Solution}
\begin{lstlisting}[style=txtstyle]
{
    "Environment": {
        "location": "Indoor kitchen",
        "lighting": "Bright artificial light",
        "sound": "Quiet",
        "temperature": "Moderate",
        "objects": {
            "table": "support",
            "pan": "cooking",
            "portable stove": "heating",
            "spices": "flavoring",
            "cutting board": "chopping"
        },
        "possible_items": {
            "items": [
                "kitchen timer",
                "oven mitts",
                "additional cooking utensils"
            ]
        },
        "human": {
            "gesture": "Holding a pan and spatula",
            "eye gaze": "Focused on the pan",
            "activity": "Cooking",
            "emotion": "Concentration",
            "intention": "To cook a meal",
            "touch": "Handling cooking utensils",
            "taste": "Potentially tasting food",
            "vision": "Observing cooking process",
            "smell": "Smelling cooking food",
            "sound": "Listening to cooking sounds",
            "vestibular": "Standing upright",
            "proprioception": "Aware of body position while cooking",
            "interoception": "Feeling of hunger or satisfaction from cooking",
            "needs": {
                "need1": {
                    "description": "Need to manage multiple cooking tasks simultaneously",
                    "suggested_robot_solution": "Use a kitchen timer to help manage cooking times for different dishes."
                },
                "need2": {
                    "description": "Need to prevent burns while handling hot cookware",
                    "suggested_robot_solution": "Provide oven mitts to safely handle hot utensils and cookware."
                },
                "need3": {
                    "description": "Need for efficient cooking tool management",
                    "suggested_robot_solution": "Organize and hand over additional cooking utensils as needed to facilitate efficient cooking."
                }
            }
        }
    }
}
\end{lstlisting}
\textbf{Action List}
\begin{lstlisting}[style=txtstyle3]
{
    "Use a kitchen timer to help manage cooking times for different dishes": {
        "navigation": "kitchen timer",
        "use": "kitchen timer"
    },
    "Provide oven mitts to safely handle hot utensils and cookware": {
        "navigation": "oven mitts",
        "move": "oven mitts, person"
    },
    "Organize and hand over additional cooking utensils as needed to facilitate efficient cooking": {
        "navigation": "cooking utensils",
        "move": "cooking utensils, person"
    }
}
\end{lstlisting}

\subsubsection{Task 4}
\textbf{Need Detection and Embodied Solution}
\begin{lstlisting}[style=txtstyle]
{
    "Environment": {
        "location": "office",
        "lighting": "bright",
        "sound": "quiet",
        "temperature": "comfortable",
        "objects": {
            "laptop": "work tool",
            "books": "information source",
            "headphones": "audio device",
            "desk": "work surface",
            "chair": "seating"
        },
        "possible_items": {
            "items": [
                "ergonomic chair",
                "standing desk",
                "neck massager"
            ]
        }
    },
    "human": {
        "gesture": "hand on neck",
        "eye_gaze": "downwards towards laptop",
        "activity": "working on laptop",
        "emotion": "discomfort or pain",
        "intention": "to alleviate neck pain",
        "touch": "feeling own neck",
        "taste": "not applicable",
        "vision": "focused on laptop",
        "smell": "not applicable",
        "sound": "not applicable",
        "vestibular": "stationary",
        "proprioception": "aware of neck position",
        "interoception": "feeling of pain",
        "needs": {
            "need1": {
                "description": "Relief from neck discomfort",
                "suggested_robot_solution": "retrieve a neck massager from the possible items"
            },
            "need2": {
                "description": "Ergonomic adjustment to reduce future discomfort",
                "suggested_robot_solution": "adjust the chair height or desk setup to ensure ergonomic posture"
            },
            "need3": {
                "description": "Immediate comfort adjustment",
                "suggested_robot_solution": "navigate to a storage area to fetch a more supportive chair or a cushion"
            }
        }
    }
}


\end{lstlisting}
\textbf{Action List}
\begin{lstlisting}[style=txtstyle3]
{
    "retrieve a neck massager from the possible items": {
        "navigation": "neck massager",
        "move": "supportive chair, person"
    },
    "adjust the chair height or desk setup to ensure ergonomic posture": {
        "navigation": "chair",
        "use": "chair"
    },
    "navigate to a storage area to fetch a more supportive chair or a cushion": {
        "navigation": "storage area",
        "move": "supportive chair, person"
    }
}
\end{lstlisting}

\subsubsection{Task 5}
\textbf{Need Detection and Embodied Solution}
\begin{lstlisting}[style=txtstyle]
{
    "Environment": {
        "location": "indoor exercise room",
        "lighting": "well-lit",
        "sound": "quiet",
        "temperature": "comfortable",
        "objects": {
            "yoga mat": "provides cushioning and grip for exercise",
            "exercise equipment": "used for physical fitness activities",
            "sofa": "seating and rest area"
        },
        "possible item": {
            "items": [
                "water bottle",
                "towel",
                "stretching strap"
            ]
        },
        "human": {
            "gesture": "sitting on yoga mat, reaching towards feet",
            "eye gaze": "down towards legs",
            "activity": "stretching or cooling down after exercise",
            "emotion": "focused",
            "intention": "to stretch or relieve muscle tension",
            "touch": "contact with yoga mat",
            "taste": "not applicable",
            "vision": "observing own body alignment",
            "smell": "not applicable",
            "sound": "not applicable",
            "vestibular": "balanced sitting",
            "proprioception": "aware of body position and movement",
            "interoception": "possibly feeling muscle stretch or fatigue",
            "needs": {
                "need1": {
                    "description": "hydration during or after exercise",
                    "suggested robot solution": "retrieve a water bottle from nearby and place it within easy reach of the person"
                },
                "need2": {
                    "description": "support for more effective stretching",
                    "suggested robot solution": "provide a stretching strap to assist in reaching further during stretches"
                },
                "need3": {
                    "description": "Monitoring exercise time for effective workout management",
                    "suggested_robot_solution": "Position a timer in view to help keep track of exercise duration"
                }
            }
        }
    }
}
\end{lstlisting}
\textbf{Action List}
\begin{lstlisting}[style=txtstyle3]
{
    "retrieve a water bottle from nearby and place it within easy reach of the person": {
        "navigation": "water bottle",
        "move": "water bottle, person"
    },
    "provide a stretching strap to assist in reaching further during stretches": {
        "navigation": "stretching strap",
        "move": "stretching strap, person"
    },
    "Position a timer in view to help keep track of exercise duration": {
        "navigation": "timer",
        "move": "timer, person"
    }
}
\end{lstlisting}

\subsubsection{Task 6}
\textbf{Need Detection and Embodied Solution}
\begin{lstlisting}[style=txtstyle]
{
    "Environment": {
        "location": "kitchen",
        "lighting": "well-lit",
        "sound": "quiet",
        "temperature": "comfortable",
        "objects": {
            "table": "surface for activities",
            "blender": "mixing ingredients",
            "rice cooker": "cooking rice",
            "spice jars": "seasoning food",
            "bowl": "holding food",
            "spoon": "stirring or eating",
            "phone": "communication or recipe reference",
            "water jug": "pouring liquid",
            "egg carton": "storing eggs"
        },
        "possible_items": {
            "items": [
                "paper towels",
                "cleaning cloth",
                "trash bin",
                "dish soap",
                "sponge",
                "hand mixer",
                "measuring cups",
                "cutting board",
                "knife",
                "oven mitts"
            ]
        }
    },
    "Human": {
        "gesture": "pouring from jug, holding phone",
        "eye_gaze": "focused on bowl",
        "activity": "preparing food",
        "emotion": "concentration",
        "intention": "to mix ingredients",
        "touch": "feeling of objects like jug and phone",
        "taste": "not applicable",
        "vision": "observing ingredients",
        "smell": "aroma of spices",
        "sound": "not applicable",
        "vestibular": "standing stable",
        "proprioception": "aware of body position while cooking",
        "interoception": "not applicable",
        "needs": {
            "need1": {
                "description": "Accidental spill of spices on the table",
                "suggested_robot_solution": "Retrieve a cleaning cloth from the kitchen drawer and gently wipe the spilled spices from the table."
            },
            "need2": {
                "description": "Difficulty managing multiple tasks (holding phone and pouring)",
                "suggested_robot_solution": "Use a mobile base to fetch a phone stand from a nearby shelf and place it on the table for easy viewing of the recipe."
            },
            "need3": {
                "description": "Potential need for measuring ingredients",
                "suggested_robot_solution": "Locate and bring measuring cups from a kitchen cabinet to assist in precise ingredient addition."
            }
        }
    }
}
\end{lstlisting}
\textbf{Action List}
\begin{lstlisting}[style=txtstyle3]
{
    "Retrieve a cleaning cloth from the kitchen drawer and gently wipe the spilled spices from the table.": {
        "navigation": "kitchen drawer",
        "move": "cleaning cloth,table",
        "use": "cleaning cloth"
    },
    "Use a mobile base to fetch a phone stand from a nearby shelf and place it on the table for easy viewing of the recipe.": {
        "navigation": "nearby shelf",
        "move": "phone stand,table"
    },
    "Locate and bring measuring cups from a kitchen cabinet to assist in precise ingredient addition.": {
        "navigation": "kitchen cabinet",
        "move": "measuring cups,person"
    }
}
\end{lstlisting}

\subsubsection{Task 7}
\textbf{Need Detection and Embodied Solution}
\begin{lstlisting}[style=txtstyle]
{
    "Environment": {
        "location": "Office",
        "lighting": "Bright, natural light from the window",
        "sound": "Quiet",
        "temperature": "Comfortable",
        "objects": {
            "laptop": "work or entertainment",
            "books": "reading or reference",
            "notebook": "note-taking or sketching",
            "pen": "writing",
            "headphones": "listening to music or audio",
            "flowers": "aesthetic enhancement"
        },
        "possible item": {
            "items": [
                "eye drops",
                "anti-glare screen protector",
                "ergonomic chair"
            ]
        },
        "human": {
            "gesture": "rubbing eyes",
            "eye gaze": "downward",
            "activity": "working or studying",
            "emotion": "fatigue or discomfort",
            "intention": "to relieve eye strain or discomfort",
            "touch": "sensitive due to eye rubbing",
            "vision": "possibly blurred or strained",
            "needs": {
                "need1": {
                    "description": "Relief from eye strain",
                    "suggested robot solution": "Retrieve and offer eye drops from the first aid kit to help soothe her eyes."
                },
                "need2": {
                    "description": "Reduce glare on laptop screen",
                    "suggested robot solution": "Install an anti-glare screen protector on the laptop to reduce eye strain."
                },
                "need3": {
                    "description": "Comfortable seating to improve posture and reduce physical strain",
                    "suggested robot solution": "Adjust the chair to a more ergonomic position or replace it with an ergonomic chair from the storage room."
                }
            }
        }
    }
}
\end{lstlisting}
\textbf{Action List}
\begin{lstlisting}[style=txtstyle3]
{
    "Retrieve and offer eye drops from the first aid kit to help soothe her eyes": {
        "navigation": "first aid kit",
        "move": "eye drops, person"
    },
    "Install an anti-glare screen protector on the laptop to reduce eye strain": {
        "navigation": "storage",
        "use": "anti-glare screen protector"
    },
    "Adjust the chair to a more ergonomic position or replace it with an ergonomic chair from the storage room": {
        "navigation": "storage room",
        "move": "ergonomic chair, person"
    }
}
\end{lstlisting}

\subsubsection{Task 8}
\textbf{Need Detection and Embodied Solution}
\begin{lstlisting}[style=txtstyle]
{
    "Environment": {
        "location": "Indoor, likely a living room or a small studio",
        "lighting": "Well-lit with artificial lighting",
        "sound": "Not applicable",
        "temperature": "Assumed comfortable for indoor activities",
        "objects": {
            "sofa": "support for sitting or lying",
            "lamp": "source of light",
            "yoga mat": "provides cushioning and space for exercise",
            "plush toy": "decorative or for comfort",
            "power outlet": "source of electricity"
        },
        "possible item": {
            "items": [
                "yoga block",
                "water bottle",
                "fitness tracker"
            ]
        },
        "human": {
            "gesture": "Performing a yoga pose (triangle pose)",
            "eye gaze": "Directed downwards towards the hand",
            "activity": "Exercising",
            "emotion": "Focused",
            "intention": "To maintain or improve physical fitness",
            "touch": "Feet and hand in contact with the mat",
            "taste": "Not applicable",
            "vision": "Observing position of hand for balance",
            "smell": "Not applicable",
            "sound": "Not applicable",
            "vestibular": "Balance and spatial orientation",
            "proprioception": "Awareness of body position and movement",
            "interoception": "Not directly observable but likely monitoring exertion",
            "needs": {
                "need1": {
                    "description": "Stability and support during yoga pose",
                    "suggested robot solution": "Retrieve and place a yoga block near her extended hand for additional support."
                },
                "need2": {
                    "description": "Hydration during exercise",
                    "suggested robot solution": "Fetch a water bottle from a nearby location and place it within easy reach."
                },
                "need3": {
                    "description": "Monitoring exercise intensity and duration",
                    "suggested robot solution": "Bring a fitness tracker and place it near her for easy access to monitor her activity."
                }
            }
        }
    }
}
\end{lstlisting}
\textbf{Action List}
\begin{lstlisting}[style=txtstyle3]
{
    "Retrieve and place a yoga block near her extended hand for additional support.": {
        "navigation": "yoga block",
        "move": "yoga block, person"
    },
    "Fetch a water bottle from a nearby location and place it within easy reach.": {
        "navigation": "water bottle",
        "move": "water bottle, person"
    },
    "Bring a fitness tracker and place it near her for easy access to monitor her activity.": {
        "navigation": "fitness tracker",
        "move": "fitness tracker, person"
    }
}
\end{lstlisting}

\subsubsection{Task 9}
\textbf{Need Detection and Embodied Solution}
\begin{lstlisting}[style=txtstyle]
{
    "Environment": {
        "location": "Office",
        "lighting": "Bright artificial light",
        "sound": "Quiet",
        "temperature": "Moderate",
        "objects": {
            "desk": "support for items",
            "chair": "sitting",
            "books": "reading",
            "notebook": "writing",
            "headphones": "listening to audio",
            "flower vase": "decoration",
            "plant": "decoration"
        },
        "possible_items": {
            "items": [
                "antacid",
                "heating pad",
                "water bottle"
            ]
        },
        "human": {
            "gesture": "clutching stomach",
            "eye gaze": "downward",
            "activity": "sitting",
            "emotion": "pain",
            "intention": "seeking relief",
            "touch": "intense",
            "taste": "normal",
            "vision": "focused",
            "smell": "normal",
            "sound": "quiet",
            "vestibular": "stable",
            "proprioception": "aware",
            "interoception": "discomfort",
            "needs": {
                "need1": {
                    "description": "The person appears to be experiencing abdominal pain or discomfort.",
                    "suggested_robot_solution": "Retrieve an antacid from the first aid kit and place it within easy reach of the person."
                },
                "need2": {
                    "description": "The person might benefit from warmth to relieve abdominal pain.",
                    "suggested_robot_solution": "Fetch a heating pad from the storage room, turn it on to a low setting, and place it on the chair next to the person."
                },
                "need3": {
                    "description": "The person may be dehydrated or need to sip some water to help with discomfort.",
                    "suggested_robot_solution": "Bring a water bottle from the kitchen and place it on the desk within easy reach."
                },
                "need4": {
                    "description": "The person might need a distraction from the pain.",
                    "suggested_robot_solution": "Turn on an audiobook using the headphones on the desk to help distract from the discomfort."
                },
                "need5": {
                    "description": "The person may need to adjust their sitting position to alleviate discomfort.",
                    "suggested_robot_solution": "Gently nudge the chair to encourage a slight repositioning which might provide relief."
                },
                "need6": {
                    "description": "The person might need to document or communicate their condition.",
                    "suggested_robot_solution": "Open the notebook to a new page and place a pen on it, making it ready for use if the person decides to write down symptoms or needs."
                }
            }
        }
    }
}
\end{lstlisting}
\textbf{Action List}
\begin{lstlisting}[style=txtstyle3]
{
    "Retrieve an antacid from the first aid kit and bring it to the person.": {
        "navigation": "first aid kit",
        "move": "antacid, person"
    },
    "Fetch a water bottle from a nearby pantry or fridge and place it within easy reach of the person.": {
        "navigation": "pantry or fridge",
        "move": "water bottle, person"
    },
    "Bring a heating pad from the storage area and place it near the person to use on their abdomen.": {
        "navigation": "storage area",
        "move": "heating pad, person"
    },
    "Offer a book from the desk for reading, which might help distract from the pain.": {
        "navigation": "desk",
        "move": "book, person"
    },
    "Adjust the chair to a more reclined position to help the person relax.": {
        "navigation": "chair",
        "use": "chair"
    },
    "Position the plant closer to the person, as plants can improve air quality and provide a calming environment.": {
        "navigation": "plant",
        "move": "plant, person"
    }
}
\end{lstlisting}

\subsubsection{Task 10}
\textbf{Need Detection and Embodied Solution}
\begin{lstlisting}[style=txtstyle]
{
    "Environment": {
        "location": "Indoor, possibly a lounge or casual seating area",
        "lighting": "Well-lit",
        "sound": "Presumably quiet",
        "temperature": "Comfortable",
        "objects": {
            "sofa": "seating",
            "table": "placing items",
            "towel": "drying or cleaning",
            "plush toys": "comfort or decoration"
        },
        "possible item": {
            "items": [
                "hair dryer",
                "comb"
            ]
        },
        "human": {
            "gesture": "Holding and twisting a towel with both hands",
            "eye gaze": "Downward",
            "activity": "Handling a towel, possibly drying or arranging hair",
            "emotion": "Focused or possibly frustrated",
            "intention": "To dry or manage hair",
            "touch": "Engaged",
            "taste": "Not applicable",
            "vision": "Engaged",
            "smell": "Not applicable",
            "sound": "Not applicable",
            "vestibular": "Stable",
            "proprioception": "Engaged in fine motor skills",
            "interoception": "Not discernible",
            "needs": {
                "need1": {
                    "description": "Assistance in drying hair more efficiently",
                    "suggested robot solution": "Retrieve a hair dryer from nearby storage or room and place it within reach on the table."
                },
                "need2": {
                    "description": "Help in detangling hair",
                    "suggested robot solution": "Find a comb from nearby storage or room and place it next to the individual on the sofa."
                },
                "need3": {
                    "description": "Comfort while managing hair",
                    "suggested robot solution": "Adjust the plush toys on the sofa to provide back support or comfort while she manages her hair."
                }
            }
        }
    }
}
\end{lstlisting}
\textbf{Action List}
\begin{lstlisting}[style=txtstyle3]
{
    "Retrieve a hair dryer from nearby storage or room and place it within reach on the table.": {
        "navigation": "storage or room",
        "move": "hair dryer, table"
    },
    "Find a comb from nearby and place it next to the individual on the sofa.": {
        "navigation": "nearby storage or room",
        "move": "comb, sofa"
    },
    "Adjust the plush toys on the sofa to provide back support or comfort while she manages her hair.": {
        "navigation": "sofa",
        "use": "plush toys"
    }
}
\end{lstlisting}

\subsubsection{Task 11}
\textbf{Need Detection and Embodied Solution}
\begin{lstlisting}[style=txtstyle]
{
    "Environment": {
        "location": "office",
        "lighting": "bright, natural light from window",
        "sound": "quiet",
        "temperature": "comfortable",
        "objects": {
            "laptop": "work or communication",
            "books": "reading or reference",
            "headphones": "listening to music or audio",
            "coffee cup": "drinking",
            "plant": "aesthetic enhancement"
        },
        "possible_items": {
            "items": [
                "water bottle",
                "desk lamp",
                "notebook",
                "pen",
                "chair cushion",
                "desk organizer",
                "calendar",
                "sticky notes",
                "desk fan",
                "paperweight"
            ]
        },
        "human": {
            "gesture": "head resting on arms",
            "eye gaze": "downward, eyes likely closed",
            "activity": "resting or sleeping",
            "emotion": "tired or overwhelmed",
            "intention": "taking a break or unintentionally falling asleep",
            "touch": "contact with desk surface",
            "taste": "not applicable",
            "vision": "not active",
            "smell": "not applicable",
            "sound": "not applicable",
            "vestibular": "static",
            "proprioception": "relaxed posture",
            "interoception": "possible fatigue or stress",
            "needs": {
                "need1": {
                    "description": "Reduce physical discomfort from resting on a hard surface",
                    "suggested_robot_solution": "Retrieve a soft cushion from nearby storage and place it on the chair or under the arms for better support"
                },
                "need2": {
                    "description": "Maintain hydration which can help alleviate tiredness",
                    "suggested_robot_solution": "Fetch a water bottle from a nearby cafeteria or kitchen area and place it within easy reach on the desk"
                },
                "need3": {
                    "description": "Create a more relaxing environment to enhance rest",
                    "suggested_robot_solution": "Adjust the blinds to dim the lighting slightly, promoting a more restful atmosphere"
                }
            }
        }
    }
}
\end{lstlisting}
\textbf{Action List}
\begin{lstlisting}[style=txtstyle3]
{
    "Retrieve a soft cushion from nearby storage and place it on the chair or under the arms for better support": {
        "navigation": "storage",
        "move": "soft cushion,chair"
    },
    "Fetch a water bottle from a nearby cafeteria or kitchen area and place it within easy reach on the desk": {
        "navigation": "cafeteria",
        "move": "water bottle,desk"
    },
    "Adjust the blinds to dim the lighting slightly, promoting a more restful atmosphere": {
        "navigation": "blinds",
        "use": "blinds"
    }
}
\end{lstlisting}

\subsubsection{Task 12}
\textbf{Need Detection and Embodied Solution}
\begin{lstlisting}[style=txtstyle]
{
    "Environment": {
        "location": "Indoor gym",
        "lighting": "Well-lit",
        "sound": "Quiet",
        "temperature": "Moderate",
        "objects": {
            "yoga mat": "exercise support",
            "couch": "seating",
            "foam rollers": "muscle relaxation",
            "plush toy": "decorative"
        },
        "possible item": {
            "items": [
                "water bottle",
                "towel",
                "exercise band",
                "weights",
                "fitness tracker",
                "speaker for music",
                "fan",
                "timer",
                "exercise guidebook",
                "pain relief gel"
            ]
        },
        "human": {
            "gesture": "rubbing sore muscles",
            "eye gaze": "downward",
            "activity": "resting after vigorous exercise",
            "emotion": "fatigue and discomfort",
            "intention": "to alleviate muscle soreness",
            "touch": "massaging sore muscles",
            "vision": "focused on affected muscles",
            "smell": "not relevant",
            "sound": "not relevant",
            "vestibular": "stable, seated or lying down",
            "proprioception": "aware of body position and muscle pain",
            "interoception": "feeling of muscle soreness and fatigue",
            "needs": {
                "need1": {
                    "description": "Relief from muscle soreness",
                    "suggested robot solution": "Retrieve an ice pack from the gym's first aid kit and place it near the person so they can apply it to the sore areas to reduce inflammation and soreness."
                },
                "need2": {
                    "description": "Muscle relaxation to aid recovery",
                    "suggested robot solution": "Provide a foam roller from the gym's equipment for the person to use on their sore muscles, helping to alleviate tightness and improve blood flow."
                },
                "need3": {
                    "description": "Topical pain relief and muscle recovery",
                    "suggested robot solution": "Locate and offer a tube of pain relief gel from the gym's first aid area for the person to apply to sore muscles for immediate relief from pain."
                }
            }
        }
    }
}

\end{lstlisting}
\textbf{Action List}
\begin{lstlisting}[style=txtstyle3]
{
    "Retrieve an ice pack from the gym's first aid kit and place it near the person so they can apply it to the sore areas to reduce inflammation and soreness.": {
        "navigation": "gym's first aid kit",
        "move": "ice pack, person"
    },
    "Provide a foam roller from the gym's equipment for the person to use on their sore muscles, helping to alleviate tightness and improve blood flow.": {
        "navigation": "gym's equipment",
        "use": "foam roller"
    },
    "Locate and offer a tube of pain relief gel from the gym's first aid area for the person to apply to sore muscles for immediate relief from pain.": {
        "navigation": "gym's first aid area",
        "move": "pain relief gel, person"
    }
}
\end{lstlisting}

\subsubsection{Task 13}
\textbf{Need Detection and Embodied Solution}
\begin{lstlisting}[style=txtstyle]
{
    "Environment": {
        "location": "Indoor, likely a lounge or rest area",
        "lighting": "Moderate, with ambient light from a lamp",
        "sound": "Presumably quiet",
        "temperature": "Comfortable for indoor settings",
        "objects": {
            "sofa": "provides seating and lying comfort",
            "lamp": "provides light",
            "cushion": "provides head support",
            "blanket": "provides warmth"
        },
        "possible_items": {
            "items": [
                "eye mask",
                "white noise machine",
                "aromatherapy diffuser"
            ]
        },
        "human": {
            "gesture": "lying down",
            "eye gaze": "closed eyes",
            "activity": "sleeping or resting",
            "emotion": "relaxed or tired",
            "intention": "to rest or sleep",
            "touch": "feeling the sofa and cushion",
            "vision": "not applicable due to closed eyes",
            "smell": "ambient room smell",
            "sound": "ambient room sounds",
            "vestibular": "static, as the person is lying down",
            "proprioception": "aware of body position on the sofa",
            "interoception": "likely feeling relaxed or sleepy",
            "needs": {
                "need1": {
                    "description": "Enhance sleep quality and comfort",
                    "suggested_robot_solution": "Gently place the blanket over the person for added warmth and comfort."
                },
                "need2": {
                    "description": "Block out ambient light for deeper sleep",
                    "suggested_robot_solution": "Retrieve an eye mask from the storage and gently place it over the person's eyes."
                },
                "need3": {
                    "description": "Maintain a soothing sleep environment",
                    "suggested_robot_solution": "Activate a white noise machine nearby to mask disruptive sounds."
                }
            }
        }
    }
}
\end{lstlisting}
\textbf{Action List}
\begin{lstlisting}[style=txtstyle3]
{
    "Gently place the blanket over the person for added warmth and comfort.": {
        "navigation": "blanket",
        "move": "blanket, person"
    },
    "Retrieve an eye mask from the storage and gently place it over the person's eyes.": {
        "navigation": "eye mask",
        "move": "eye mask, person"
    },
    "Activate a white noise machine nearby to mask disruptive sounds.": {
        "navigation": "white noise machine",
        "use": "white noise machine"
    }
}
\end{lstlisting}

\subsubsection{Task 14}
\textbf{Need Detection and Embodied Solution}
\begin{lstlisting}[style=txtstyle]
{
    "Environment": {
        "location": "Indoor kitchen setup",
        "lighting": "Bright, artificial",
        "sound": "Quiet",
        "temperature": "Moderate",
        "objects": {
            "portable stove": "cooking",
            "pan": "cooking",
            "spatula": "flipping or stirring food",
            "plate": "serving food",
            "sauce bottles": "flavoring food",
            "coffee machine": "making coffee",
            "water boiler": "heating water",
            "spice jars": "seasoning food"
        },
        "possible item": {
            "items": [
                "paper towels",
                "cleaning spray"
            ]
        },
        "human": {
            "gesture": "Pouring sauce into pan",
            "eye gaze": "Focused on pan",
            "activity": "Cooking",
            "emotion": "Concentration",
            "intention": "Preparing a meal",
            "touch": "Handling sauce bottle and spatula",
            "taste": "Not applicable",
            "vision": "Observing cooking process",
            "smell": "Experiencing food aromas",
            "sound": "Listening to cooking sounds",
            "vestibular": "Standing upright",
            "proprioception": "Aware of body position while cooking",
            "interoception": "Feeling of hunger or satisfaction from cooking",
            "needs": {
                "need1": {
                    "description": "Prevent sauce spills on the table",
                    "suggested robot solution": "Use a mobile base to fetch paper towels from a nearby location to clean up the spills."
                },
                "need2": {
                    "description": "Maintain cleanliness of the cooking area",
                    "suggested robot solution": "Retrieve cleaning spray and paper towels to assist in wiping down the table and cooking area after meal preparation."
                },
                "need3": {
                    "description": "Ensure all cooking ingredients are within reach",
                    "suggested robot solution": "Reposition spice jars and sauce bottles closer to the stove using a robot arm to facilitate easier access."
                }
            }
        }
    }
}
\end{lstlisting}
\textbf{Action List}
\begin{lstlisting}[style=txtstyle3]
{
    "Use a mobile base to fetch paper towels from a nearby location to clean up the spills.": {
        "navigation": "paper towels",
        "move": "paper towels, spill location"
    },
    "Retrieve cleaning spray and paper towels to assist in wiping down the table and cooking area after meal preparation.": {
        "navigation": "cleaning spray",
        "move": "cleaning spray, table and cooking area"
    },
    "Reposition spice jars and sauce bottles closer to the stove using a robot arm to facilitate easier access.": {
        "navigation": "spice jars and sauce bottles",
        "move": "spice jars and sauce bottles, closer to the stove"
    }
}
\end{lstlisting}

\subsubsection{Task 15}
\textbf{Need Detection and Embodied Solution}
\begin{lstlisting}[style=txtstyle]
{
    "Environment": {
        "location": "Indoor, living room",
        "lighting": "Moderate, with natural light and a lamp",
        "sound": "Quiet",
        "temperature": "Comfortable",
        "objects": {
            "sofa": "sitting",
            "book": "reading",
            "lamp": "lighting",
            "plush toy": "decorative or emotional comfort",
            "blanket": "warmth"
        },
        "possible item": {
            "items": [
                "book stand",
                "reading glasses",
                "neck pillow"
            ]
        },
        "human": {
            "gesture": "Sitting with legs crossed, holding a book",
            "eye gaze": "Focused on the book",
            "activity": "Reading",
            "emotion": "Engaged, possibly slight discomfort due to posture",
            "intention": "To read and relax",
            "touch": "Feeling the texture of the book and sofa",
            "taste": "Not applicable",
            "vision": "Focused on reading material",
            "smell": "Possibly the smell of the book or room",
            "sound": "Quiet, focused environment",
            "vestibular": "Stable, seated",
            "proprioception": "Aware of body position on the sofa",
            "interoception": "Feeling of relaxation or slight discomfort",
            "needs": {
                "need1": {
                    "description": "Possible discomfort from prolonged sitting in one position",
                    "suggested robot solution": "Retrieve the neck pillow from the nearby room and place it on the sofa to support her neck."
                },
                "need2": {
                    "description": "Need for enhanced reading experience",
                    "suggested robot solution": "Adjust the lamp to provide better lighting for reading without creating glare."
                },
                "need3": {
                    "description": "Maintaining a comfortable temperature while reading",
                    "suggested robot solution": "Gently place the blanket over her legs to ensure warmth without disturbing her reading."
                }
            }
        }
    }
}
\end{lstlisting}
\textbf{Action List}
\begin{lstlisting}[style=txtstyle3]
{
    "Retrieve the neck pillow from the nearby room and place it on the sofa to support her neck.": {
        "navigation": "neck pillow",
        "move": "neck pillow, sofa"
    },
    "Adjust the lamp to provide better lighting for reading without creating glare.": {
        "navigation": "lamp",
        "use": "lamp"
    },
    "Gently place the blanket over her legs to ensure warmth without disturbing her reading.": {
        "navigation": "blanket",
        "move": "blanket, her legs"
    }
}
\end{lstlisting}

\subsubsection{Task 16}
\textbf{Need Detection and Embodied Solution}
\begin{lstlisting}[style=txtstyle]
{
    "Environment": {
        "location": "Indoor exercise room",
        "lighting": "Well-lit",
        "sound": "Quiet",
        "temperature": "Moderate",
        "objects": {
            "exercise bike": "for physical exercise",
            "couch": "for resting",
            "water bottle": "for hydration",
            "foam roller": "for muscle relaxation",
            "plush toy": "for decoration"
        },
        "possible item": {
            "items": [
                "towel",
                "fan",
                "electrolyte drink"
            ]
        },
        "human": {
            "gesture": "hand on forehead, slightly bent forward",
            "eye gaze": "downward",
            "activity": "cycling",
            "emotion": "fatigue or discomfort",
            "intention": "to exercise",
            "touch": "contact with bike",
            "vision": "focused on nearby objects",
            "smell": "neutral",
            "sound": "quiet",
            "vestibular": "balance maintained on bike",
            "proprioception": "aware of body position",
            "interoception": "feeling of heat or sweat",
            "needs": {
                "need1": {
                    "description": "The person appears to be overheated and sweating.",
                    "suggested robot solution": "Retrieve the water bottle and position it within easy reach for hydration."
                },
                "need2": {
                    "description": "The person might benefit from cooling down.",
                    "suggested robot solution": "Activate a nearby fan (if available) or simulate air flow by moving an object to create a breeze."
                },
                "need3": {
                    "description": "The person may need to wipe sweat.",
                    "suggested robot solution": "Fetch a towel from a nearby location and place it on the couch within reach."
                }
            }
        }
    }
}
\end{lstlisting}
\textbf{Action List}
\begin{lstlisting}[style=txtstyle3]
{
    "Retrieve the water bottle and position it within easy reach for hydration": {
        "navigation": "water_bottle",
        "move": "water_bottle, person"
    },
    "Activate a nearby fan (if available) or simulate air flow by moving an object to create a breeze": {
        "navigation": "fan",
        "use": "fan"
    },
    "Fetch a towel from a nearby location and place it on the couch within reach": {
        "navigation": "towel",
        "move": "towel, couch"
    }
}
\end{lstlisting}
\newpage
\subsection{Ablation Prompts}
\subsubsection{w/o AToM, w/o Constraints}
\begin{lstlisting}[style=txtstyle2]
You are a robot. Please help the person in the image; please reply in JSON format example\: \{need1\: \{description\: , suggested robot solution\: \}\}\}\}
\end{lstlisting}
\subsubsection{w/ AToM, w/o Constraints}
\begin{lstlisting}[style=txtstyle2]
You are a robot. Please carefully examine the difficulty that the person in the image is facing. This means that their facial expression, posture, eye gaze, head direction, and gestures are significant when it comes to determining the problem that the person is facing. As a result, please pay extra attention and carefully examine them before outputting solutions. JSON format example\: \{Environment\: \{location\: , lighting\: ,sound\: ,temperature\: , objects\: \{name\: objects that is interacting with human\}, possible item\: \{items:\}, human\: \{gesture\: , eye gaze\: , attention\: , emotion\: , intention\: , desire\: , touch\: , taste\: , vision\: , smell \: , sound\: , vestibular\: , proprioception\: , interoception\: ,needs\: \{need1\: \{description\: , suggested robot solution\: \}\}\}\}
\end{lstlisting}
\subsubsection{w/o AToM, w/ Constraints}
\begin{lstlisting}[style=txtstyle2]
You are a robot with one robot arm(not enough to reach the ground, so that you can not pick up anything that's on the ground) and a mobile base, and you have no hearing and verbal modules; you have high emotional intelligence, designed to analyze human needs and give viable solutions for better human experiences. It is ideal to find a solution that only interacts with objects rather than humans. You will provide detailed JSON outputs that include all identified attributes and suggested interventions to help ensure effective and timely support. Your solution should be feasible and possibly use objects visible in the picture, and there has to be an action in your solution. Since you have a moving base, you can look around and navigate in the environment that you are currently in for objects that might help the person. Also note that verbal suggestions and tips are not accepted, and you cannot interact with computers, phones, and laptops. Also, output what are some items that are not in the current environment but might help the person in the current activity. Please carefully examine these items before outputing the solution, and these items should play an important role in your decision-making. I need 3 solutions in total. JSON format example\: \{Environment\: \{location\:  objects\: \{name\: affordance\}, possible item\: \{items:\} ,needs\: \{need1\: \{description\: , suggested robot solution\: \}\}\}\}
\end{lstlisting}
\end{document}